\documentclass{article}
\usepackage{ijcai19}

\usepackage{times}
\usepackage{soul}
\usepackage{url}
\usepackage[hidelinks]{hyperref}
\usepackage[utf8]{inputenc}
\usepackage[small]{caption}
\usepackage{graphicx}
\usepackage{amsmath}
\usepackage{booktabs}
\usepackage{algorithm}
\usepackage{algorithmic}
\usepackage{epsfig}
\usepackage{amssymb}
\usepackage{multirow}
\usepackage{float}
\usepackage{dblfloatfix}

\DeclareMathOperator*{\argmin}{arg\,min}

\title{Play and Prune: Adaptive Filter Pruning for Deep Model Compression}

\author{
Pravendra Singh$^1$\and
Vinay Kumar Verma$^1$\and
Piyush Rai$^1$\And
Vinay P. Namboodiri$^1$
\affiliations
$^1$Department of Computer Science and Engineering, IIT Kanpur, India\\
\emails
\{psingh, vkverma, piyush, vinaypn\}@cse.iitk.ac.in
}

\begin{document}

\maketitle

\begin{abstract}
  While convolutional neural networks (CNN) have achieved impressive performance on various classification/recognition tasks, they typically consist of a massive number of parameters. This results in significant memory requirement as well as computational overheads. Consequently, there is a growing need for filter-level pruning approaches for compressing CNN based models that not only reduce the total number of parameters but reduce the overall computation as well. We present a new min-max framework for filter-level pruning of CNNs. Our framework, called Play and Prune (PP), jointly prunes and fine-tunes CNN model parameters, with an adaptive pruning rate, while maintaining the model's predictive performance. Our framework consists of two modules: (1) An adaptive filter pruning (AFP) module, which minimizes the number of filters in the model; and (2) A pruning rate controller (PRC) module, which maximizes the accuracy during pruning. Moreover, unlike most previous approaches, our approach allows directly specifying the desired error tolerance instead of pruning level. Our compressed models can be deployed at run-time, without requiring any special libraries or hardware. Our approach reduces the number of parameters of VGG-16 by an impressive factor of 17.5X, and number of FLOPS by 6.43X, with no loss of accuracy, significantly outperforming other state-of-the-art filter pruning methods.
\end{abstract}
\vspace{-1em}
\section{Introduction}
Deep convolutional neural networks (CNN) have been used widely for object recognition and various other computer vision tasks. After the early works based on standard forms of deep convolutional neural networks \cite{lecun1998gradient,krizhevsky2012imagenet}, recent works have proposed and investigated various architectural changes \cite{vgg2014very,chollet2017xception,resnet} to improve the performance of CNNs. Although these changes, such as adding more layers to the CNN, have led to impressive performance gains, and they have also resulted in a substantial increase in the number of parameters, as well as the computational cost. The increase in model size and computations have made it impractical to deploy these models on embedded and mobile devices for real-world applications. To address this, recent efforts have focused on several approaches for compressing CNNs, such as using binary or quantized \cite{binarycompression} weights. However, these require special hardware or pruning of unimportant/redundant weights \cite{han2015deep,han2015nips,alvarez2016learning,wen2016nips_sparse}, and give rather limited speedups.

As most of the CNN parameters reside in the fully connected layers, a high compression rate with respect to the number of network parameters can be achieved by simply pruning redundant neurons from the fully connected layers. However, this does not typically result in any significant reduction in computations (FLOPs based speedup), as most of the computations are performed in convolutional layers. For example, in case of  VGG-16, the fully connected layers contain 90\% of total parameters but account for only 1\% of computations, which means that convolutional layers, despite having only about 10\% of the total parameters, are responsible for 99\% of computations.  This has led to considerable recent interest in convolutional layer filter pruning approaches. However, most existing pruning approaches \cite{han2015deep,han2015nips}  result in irregular sparsity in the convolutional filters, which requires software specifically designed for sparse tensors to achieve speed-ups in practice \cite{han2015deep}. In contrast, some other filter pruning approaches \cite{luo2017thinet,weightSumICLR-17,channelPruning17} are designed to directly reduce the feature map width by removing specific convolutional filters via $\ell_2$ or $\ell_1$ regularization on the filters, and effectively reducing the computation, memory, and the number of model parameters. These methods result in models that can be directly used without requiring any sparse libraries or special hardware. 

In this work, we propose a novel filter pruning formulation. Our formulation is based on a simple min-max game between two modules to achieve an adaptive maximum pruning with minimal accuracy drop. We show that our approach, dubbed as Play-and-Prune (PP), results in substantially improved performance as compared to other recently proposed filter pruning strategies while being highly stable and efficient to train. We refer to the two modules of our framework as an Adaptive Filter Pruning (AFP) and Pruning Rate Controller (PRC). The AFP is responsible for pruning the convolutional filter, while the PRC is responsible for maintaining accuracy. 

Unlike most previous approaches, our approach does not require any \emph{external} fine-tuning. In each epoch, it performs an adaptive fine-tuning to recover from the accuracy loss caused by the previous epoch's pruning. Moreover, while previous approaches need to pre-specify a pruning level for each layer, our approach is more flexible in the sense that it directly specifies an error tolerance level and, based on that, decides which filters to prune, and from which layer(s). 

Through an extensive set of experiments and ablation studies on several benchmarks, we show that Play-and-Prune provides state-of-the-art filter pruning, and significantly outperforms existing methods. 

\section{Related Work}
Among one of the earliest efforts on compressing CNNs by pruning unimportant/redundant weights,   \cite{binarycompression,hashing} includes binarizing/quantizing the network weights, which reduces the computation time as well as storage requirement. The disadvantage of this approach is that it requires special hardware to run the deployed model. Transfer learning-based methods have also been used for model compression. One such approach is by \cite{hinton2015distilling,transfer2014deep} which transfers/distills the knowledge of a massive-sized network to a much smaller network. Another popular approach is to use a \emph{sparsity} constraints on the neuron's weights. These approaches \cite{zhou2016less,han2015nips} learn sparse network weights, where most of the weights are zero, and consequently can lead to very small model sizes. However, despite sparsity of the weights, the pruned models are still not computationally efficient at run-time. Moreover, these models require special library/hardware for sparse matrix multiplication because activation/feature maps are still dense, which hinders practical utility.

Most of the popular approaches that focus on model compression are based on sparsifying the fully connected layers since, typically, about 90\% of the network parameters are in the fully connected layers. However, note that the bulk of the computations take place in the convolutional layers, and consequently, these approaches do not result in computational acceleration. Only a few recent works have had the same focus as our work, i.e., on filter pruning \cite{weightSumICLR-17,luo2017thinet,channelPruning17,hu2016network}, that can be practically useful. In \cite{weightSumICLR-17}, the authors proposed filter pruning by ranking filters based on their sum of absolute weights.  They assumed that if the sum of absolute weights is sufficiently small, the corresponding activation map will be weak. Similarly, \cite{luo2017entropy} use a different approach to rank the filter importance, based on the entropy measures. The assumption is that high entropy filters are more important. Alternatively, \cite{hu2016network} use a data-driven approach to calculate filter importance, that is based on the average percentage of zeros in the corresponding activation map. Less important filters have more number of zeros in their activation map. Recently, \cite{molchanov2016pruning} proposed improving run time by using a Taylor approximation. This approach estimates the change in cost by pruning the filters. Another work \cite{luo2017thinet} uses pruning of filters based on the next layer statistics. Their approach is based on checking the activation map of the next layer to prune the convolution filters from the current layer. In a recent work \cite{channelPruning17} used a similar approach as in \cite{luo2017thinet} but used lasso regression.

\begin{figure}[t]
    \centering
    \includegraphics[scale=0.24]{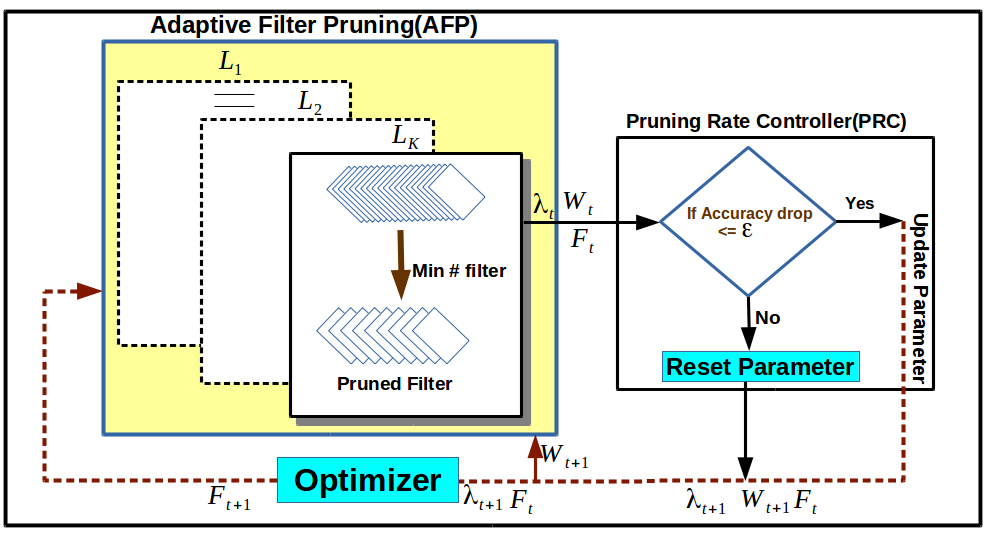}
    \caption{The figure shows the complete architecture. Here AFP minimizes the number of filter in model while PRC maximizes the accuracy during pruning. Here $\lambda_t,\mathbf{W_t}$ and $\mathbf{F_t}$ are the regularization parameter, weight-threshold and remaining filters in the model respectively at $t^{th}$ pruning iteration.}
    \label{fig:my_label}
    \vskip -1em
\end{figure}
\vspace{-1em}
\section{Proposed approach}
We assume we have a CNN model $\mathcal{M}$ with $K$ convolutional layer. Layer $i$ is denoted as $\mathcal{L}_i$ and consists of $n_i$ filters denoted as $F_{\mathcal{L}_i}=\{f_1.f_2.\dots,f_{n_i}\}$.
We assume that the unpruned model $\mathcal{M}$ has the accuracy of $\mathcal{E}$ and, post-pruning, the error tolerance limit is $\epsilon$.

\subsection{Overview}

Our deep model compression framework is modeled as a min-max game between two modules, Adaptive Filter Pruning (AFP) and Pruning Rate Controller (PRC).  The objective of the AFP is to iteratively minimize the number of filters in the model, while PRC iteratively tries to maximize the accuracy with the set of filters retained by AFP. The AFP will prune the filter only when the accuracy \emph{drop} is within the tolerance limit ($\epsilon$). If accuracy drop is more than $\epsilon$ then pruning stops and the PRC tries to recover the accuracy by fine-tuning the model. If PRC's fine-tuning is unable to bring the error within the tolerance level $\epsilon$, the AFP will not prune the filter from the model and game converges.

Let us denote the AFP by $\mathcal{P}$ and the PRC by $\mathcal{C}$. Our objective function can be defined as follows:
\begin{equation}\label{eq:minmax}
\small
\max\limits_{\#w}
\mathcal{C}\left( \min\limits_{\#w=\sum_{i=1}^{K}{n_i}} \mathcal{P}\left(F_{\mathcal{L}_1},F_{\mathcal{L}_2},\dots F_{\mathcal{L}_K}       \right)\right) 
\end{equation}

As shown in the above objective, the AFP ($\mathcal{P}$) minimizes the number of filters in the network, and the PRC $(\mathcal{C}$) optimizes the accuracy given that the number of filters. Here $\#w$ is the number of remaining filters after pruning by AFP.

An especially appealing aspect of our approach is that the pruning rates in each iteration are decided \emph{adaptively} based on the performance of the model. After each pruning step, the controller $\mathcal{C}$ checks the accuracy drop (see Fig.~\ref{fig:my_label}). If the accuracy drop is more than $\epsilon$, then the pruning rate is reset to zero, and the controller $\mathcal{C}$ tries to recover the system performance (further details of this part are provided in the section on PRC). Eq.~\ref{eq:minmax} converges when $\mathcal{C}(\#w)$ performance drop is more than the tolerance limit, and it is unable to recover it. In such a case, we rollback the current pruning and restore the previous model. At this point, we conclude that this is an optimal model that has the maximal filter pruning within $\epsilon$ accuracy drop.

\subsection{Convolutional Filter Partitioning}
The pruning module $\mathcal{P}$ first needs to identify a candidate set of filters to be pruned. For this, we use a filter partitioning scheme in each epoch. Suppose the entire set of filters of the model $\mathcal{M}$ is partitioned into two sets, one of which contains the important filters while the other contains the unimportant filters. Let $\mathcal{U}$ and $\mathcal{I}$ be the set of unimportant and important filters, respectively, where
\begin{equation}
    \mathcal{M}=\mathcal{U} \cup \mathcal{I} \quad \text{and} \quad \mathcal{U} \cap \mathcal{I}=\emptyset
\end{equation}
\begin{align*}
    & \mathcal{U}=\{U_{\mathcal{L}_1},U_{\mathcal{L}_2},\dots,U_{\mathcal{L}_K}\}
    \ \text{and} \ \mathcal{I}=\{I_{\mathcal{L}_1},I_{\mathcal{L}_2},\dots,I_{\mathcal{L}_K}\}
\end{align*}
Here $U_{\mathcal{L}_i}$ and $I_{\mathcal{L}_i}$ are set of unimportant and important  filters, respectively, in layer $\mathcal{L}_i$. $U_{\mathcal{L}_i}$, selected as follows:
\begin{equation}\label{eq:sow}
    U_{\mathcal{L}_i}=\underset{\text{top $\alpha$\%}}{\sigma}\left( \text{sort}(\{|f_1|,|f_2|,\dots,|f_{n_i}|\})\right)
\end{equation}
Eq.~\ref{eq:sow} sorts the set in increasing order of $|f_j|$, $\sigma$ is the select operator and selects the $\alpha\%$ filters with least importance. The remaining filters on $\mathcal{L}_i$ belongs to set $I_{\mathcal{L}_i}$. Here $|f_j|$ is the sum of absolute values of weights in convolutional filter $f_j$ and can be seen as the filter importance. A small sum of absolute values of filter coefficients implies less importance. 
Our approach to calculate filter importance uses their $\ell_1$ norm (Eq.~\ref{eq:sow}), which has been well-analyzed and used in prior works \cite{weightSumICLR-17,ding2018auto,he2018soft}. Our approach isn't however tied to this criterion, and other criteria can be used, too. We are using this criterion because of its simplicity.

\subsection{Weight Threshold Initialization}\label{sec:weightthres}
After obtaining the two sets of filters $\mathcal{U}$ and $\mathcal{I}$, directly removing $\mathcal{U}$ may result in a sharp and potentially irrecoverable accuracy drop. Therefore we only treat $\mathcal{U}$ as a candidate set of filters to be pruned, of which a \emph{subset} will be pruned eventually. To this end, we optimize the original cost function for the CNN, subject to a group sparse penalty on the set of filters in $\mathcal{U}$, as shown in Eq.~\ref{eq:l1reg}. Let $C(\Theta)$ be the original cost function, with $\Theta$ being original model parameters. The new objective function can be defined as:
\begin{equation}\label{eq:l1reg}
\Theta =\argmin\limits_\Theta \left( C(\Theta)+\lambda_A ||\mathcal{U}||_{1} \right)
\end{equation}
Here $\lambda_A$ is the $\ell_1$ regularization constant. This optimization penalizes $\mathcal{U}$ such that $|f_j|$ (sum of absolute weights of coefficients in each filter $f_j$) tends to zero, where $f_j \in U_{\mathcal{L}_i}$ $\forall i\in\{{1,2,\dots,K}\}$. This optimization also helps to transfer the information from $\mathcal{U}$ to the rest of the model. If a filter $f_j$ has approximately zero sum of absolute weights then it is deemed safe to be pruned. However, reaching a close-to-zero sum of absolute weights for the whole filter may require several epochs. We therefore choose an \emph{adaptive} weight threshold ($W_{\gamma_i}$) for each layer $\mathcal{L}_i$, such that removing 
$\forall f_j \in \mathcal{U}_{\mathcal{L}_i} \text{s.t.}  |f_j| \leq W_{\gamma_i}$ results in negligible (close to 0) accuracy drop. 

\begin{figure}[tb!]
    \centering
    \includegraphics[scale=0.31]{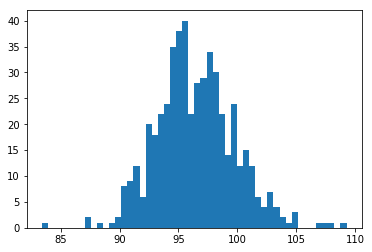}
    \includegraphics[scale=0.31]{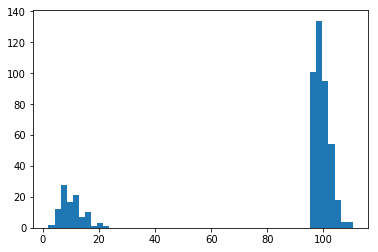}
    \vspace{-10pt}
    \caption{Histogram of the sum of absolute value of convolutional filters for CONV5\_1 in VGG-16 on CIFAR-10. Where left plot is for original filters and the right plot shows the sum of the absolute value of filters after optimization.}
     \vspace{-12pt}
    \label{fig:before_after}
\end{figure}

We calculate the initial weight threshold ($W_{\gamma_i}$) for $\mathcal{L}_i$ as follows: optimize Eq.~\ref{eq:l1reg} for one epoch with $\lambda_A = \lambda$, where $\lambda$ is the initial regularization constant, which creates two clusters of filters (if we take the sum of the absolute value of filters) as shown in Fig~\ref{fig:before_after}. On left cluster (right plot) using the binary search find the maximum threshold $W_{\gamma_i}$ for $\mathcal{L}_i$ such that accuracy drop is nearly zero. 

\subsection{Adaptive Filter Pruning (AFP)}
The objective of the AFP module is to minimize the number of filters in the model. Initially, based on the sparsity hyperparameter $\lambda$, we calculate the weight thresholds $\mathbf{W}$. Now instead of using the constant pruning rate, we change the pruning rate adaptively given by the pruning rate controller (PRC; described in the next section) in each epoch. This adaptive strategy helps to discard the filter in a balanced way, such that we can recover from the accuracy drop. In each epoch, from the current model, we select $\alpha\%$ of the filter of lowest importance from each layer, partition them into $\mathcal{U} $ and $\mathcal{I}$, and perform optimization using Eq.~\ref{eq:l1reg}, where $\lambda_A$ is given by PRC. The optimization in Eq.~\ref{eq:l1reg} transfers the knowledge of unimportant filters into the rest of the network. Therefore some filter from the $\mathcal{U}$ can be safely discarded. This removal of the filter from the model is done based on the threshold ($\mathbf{W_A}$) given by the PRC module. Now, from each layer, the filters below the adaptive threshold $\mathbf{W_A}$ are discarded. In each epoch, the weight thresholds and regularization constant is updated dynamically by the PRC module, and a subset of $\mathcal{U}$ is pruned. Hence, in the same epoch, we can recover from the accuracy drop from the previous epoch's pruning, making the model ready to prune filters in the current epoch.

The objective of the AFP module can be defined as:
\begin{equation}\label{eq:afp}
\small
\Theta'= \underset{{\#w \in \Theta'}}{\operatorname{\sigma}}\left[ \mathcal{P}\left(\argmin\limits_{\Theta'} \left( C(\Theta')+\lambda_A \sum_{i=1}^K ||\mathcal{U}||_{1} \right)\right)\right]
\end{equation}

Here $\Theta'$ is the collection of remaining filters after pruning, and $\sigma$ is the select operator. $\#w$ is the collection of all the filter from each layer $\mathcal{L}_i$ that has a sum of absolute value greater than the $W_{\gamma_i}$. From Eq.-\ref{eq:afp}, it is clear that it minimizes the number of the filters based on $W_{\gamma_i}\in \mathbf{W_A}, \forall i\in\{{1,2,\dots,K}\}$.

\subsection{Pruning Rate Controller (PRC)}
Let $\mathbf{W}=[W_{\gamma_1},W_{\gamma_2},\dots,W_{\gamma_K}]$ denote the initial weight thresholds for the $K$ layers (described in Weight Threshold Initialization section). Now the adaptive thresholds $\mathbf{W_A}$ are calculated as follows:
\begin{equation}\label{eq:weightthress}
    \mathbf{W_A}=\delta_w\times T_r\times \mathbf{W}
\end{equation}
\begin{equation}\label{eq:tolerance}
    \small
    T_r=
    \begin{cases} 
        \mathcal{C}(\#w)-(\mathcal{E}-\epsilon) &  :\mathcal{C}(\#w)-(\mathcal{E}-\epsilon) > 0 \\
        0 & :Otherwise 
   \end{cases}
\end{equation}
where $\mathcal{C}(\#w)$ is the accuracy with $\#w$ remaining filters, $\mathcal{E}$ is the accuracy of the unpruned network, and the number $\mathcal{C}(\#w)-(\mathcal{E}-\epsilon)$ denotes how far we are from tolerance error level $\epsilon$. Here, $\delta_w$ is a constant used to accelerate or decrease the pruning rate. The regularization constant $\lambda_A$ in Eq.~\ref{eq:l1reg} also adapted based on the model performance after pruning and its updates are given as follows
\begin{equation}\label{eq:lamda_tolerance}
    \small
    \mathbf{\mathbf{\lambda_{A}}}=
    \begin{cases} 
        (\mathcal{C}(\#w)-(\mathcal{E}-\epsilon))\times\lambda &  :\mathcal{C}(\#w)-(\mathcal{E}-\epsilon) > 0 \\
        0 & :Otherwise 
   \end{cases}
\end{equation}
Form Eq.~\ref{eq:lamda_tolerance} it is clear that we set the regularizer constant to zero if our pruned model performance is below the tolerance limit. Otherwise, it is proportional to the accuracy above the tolerance limit. $\lambda$ is the initial regularization constant.

The PRC module essentially controls the rate at which the filters will get pruned. In our experiments, we found that if the pruning rate is high, there is a sharp drop in accuracy after pruning, which may or may not be recoverable. Therefore pruning saturates early, and we are unable to get the high pruning rate. Also if the pruning rate is too slow, the model may get pruned very rarely and spends most of its time in fine-tuning. We, therefore, use a pruning strategy that adapts the pruning rate \emph{dynamically}. In the pruning process, if in some epoch, the system performance is below the tolerance limit, we reset the pruning rate to zero. Therefore the optimization will focus only on the accuracy gain until the accuracy is recovered to be again within the tolerance level $\epsilon$. Note that the adaptive pruning rate depends on model performance. When the model performance is within $\epsilon$, the pruning depends on how far we are from $\epsilon$. From Eq~\ref{eq:weightthress}, it is clear that the $\mathbf{W_A}$ depends on the performance of the system over the $\#w$ filters in the model. 
In this way, by controlling the pruning rate, we maintain a balance between filter pruning and accuracy. This module tries to maximize accuracy by reducing pruning rate. The objective function of PRC can be defined as:
\begin{equation}\label{eq:prc}
\max\limits_{\Theta'} \mathcal{C}\left(\Theta',D\right)
\end{equation}
Here $\mathcal{C}$ calculates the performance, i.e., accuracy. It is the function of all the convolutional filters $\Theta'$ that remain after pruning, and $D$ is the validation set used to compute the model accuracy. 

In addition to dynamically controlling the pruning rate, the PRC offers several other benefits, discussed next.

\subsubsection{Iterative Pruning Bounds Accuracy Drop}
Eq.~\ref{eq:tolerance} and Eq.~\ref{eq:lamda_tolerance} ensure that compressed model will not go beyond error tolerance limit, which is controlled by the PRC. Experimentally we found that, in a non-iterative one round pruning, if model suffers from a high accuracy drop during pruning then pruning saturates early and fine tuning will not recover accuracy drop properly. We have shown an ablation study which shows the effectiveness of iterative pruning over the single round pruning to justify this fact.
\subsubsection{Pruning Cost}
The cost/effort involved in pruning is mostly neglected in most of the existing filter pruning methods.
Moreover, most of the methods perform pruning and fine-tuning separately. In contrast, we jointly prune and fine-tune the CNN model parameters, with an adaptive pruning rate, while maintaining the model’s predictive performance. Therefore, in a given epoch, we can recover from the accuracy drop suffered due to the previous epoch’s pruning and making the model ready to prune filters in the current epoch. 

\subsubsection{Layer Importance}
Most previous methods \cite{ding2018auto,channelPruning17,weightSumICLR-17,he2018soft,yu2017nisp} use user-specified desired model compression rate but finding optimal compression rate is not so easy and involves many trials. In CNNs, some layers are relatively less important, and therefore we can prune many more filters from such layers. In contrast, if we prune a large number of filters from important layers, then this might result in an irrecoverable loss in accuracy. Our approach is more flexible since it directly specifies an error tolerance level $\epsilon$ and, based on that, adaptively decides which filters to prune, and from which layer(s), using Eq.~\ref{eq:weightthress} to determine layer-specific pruning rates.

\begin{table}[!t]
\vspace{-1em}
    \centering
    \scalebox{0.72}{
    \addtolength{\tabcolsep}{2pt}
    \begin{tabular}{|c|c|c|c|c|}
            \hline
            \multicolumn{2}{|c|}{} & \textbf{{Baseline}} & \textbf{PP-1} & \textbf{PP-2} \\ \hline
            \multicolumn{2}{|c|}{\textbf{Input Size}} & \textbf{32$\times$32$\times$3} & \textbf{32$\times$32$\times$3} & \textbf{32$\times$32$\times$3}  \\ \hline
            \multirow{16}{*}{Layers} & { CONV1\_1} &  64 &  18 &  18  \\ \cline{2-5} 
            & { CONV1\_2} & 64 &  48 &  48  \\ \cline{2-5} 
            & { CONV2\_1} & 128 & 65 &  65  \\ \cline{2-5} 
            & { CONV2\_2} & 128 & 65 &  65 \\ \cline{2-5}
            & { CONV3\_1} & 256 & 104 &  96 \\ \cline{2-5} 
            & { CONV3\_2} & 256 & 112 &  112 \\ \cline{2-5} 
            & { CONV3\_3} & 256 & 114 &  110 \\ \cline{2-5} 
            & { CONV4\_1} & 512 & 207 &  186  \\ \cline{2-5} 
            & { CONV4\_2} & 512 & 163 &  79 \\ \cline{2-5} 
            & { CONV4\_3} & 512 & 79 & 79 \\ \cline{2-5} 
            & { CONV5\_1} & 512 & 74 & 74 \\ \cline{2-5}
            & { CONV5\_2} & 512 & 48 & 48 \\ \cline{2-5}
            & { CONV5\_3} & 512 & 60 &  60 \\ \cline{2-5}
            & {FC6} & 512 & 512 & 512  \\ \cline{2-5} 
            & {FC7} & 10 & 10 & 10 \\ \hline
            \multicolumn{2}{|c|}{\textbf{Total parameters}} & 15.0M & 1.13M (13.3$\times$)  & 0.86M (17.5$\times$)  \\ \hline
            \multicolumn{2}{|c|}{\textbf{Model Size}} & 60.0 MB & 4.6 MB (13.0$\times$) & 3.5 MB (17.1$\times$)\\ \hline
            \multicolumn{2}{|c|}{\textbf{Accuracy}} & 93.49 & 93.46 & 93.35 \\ \hline
            \multicolumn{2}{|c|}{\textbf{FLOPs}} & 313.7M & 54.0M (5.8$\times$) & 48.8M (6.43$\times$) \\ \hline
        \end{tabular}}
        \vspace{-2pt}
        \caption{\small{Layer-wise pruning results and pruned models (PP-1, and PP-2) statistics for VGG-16 on CIFAR-10.}}
        \label{tabvgglike}
        \vspace{-1em}
    \end{table}

\section{Experiments and Results}

To show the effectiveness of the proposed approach, we have conducted extensive experiments on small as well as large datasets, CIFAR-10 \cite{cifar10} and ILSVRC-2012 \cite{imagenet2015}. Our approach yields state-of-art results on VGG-16 \cite{vgg2014very}, and RESNET-50 \cite{resnet} respectively.
 
For all our experiments, we set $\lambda=0.0005$ (set initially but later adapted), $\delta_w=1$ and $\alpha=10\%$. We follow the same parameter settings and training schedule as \cite{weightSumICLR-17,channelPruning17,he2018soft}. We also report an ablation study for various values of $\alpha$.  

\begin{figure}[!h]
    \centering
    \includegraphics[scale=0.27]{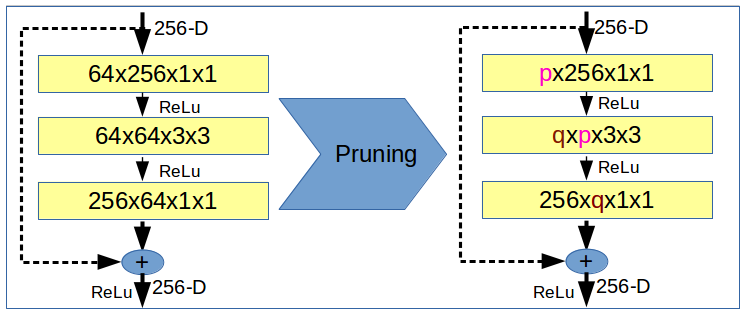}
    \caption{Our ResNet pruning strategy, where we pruned only first two convolutional layers in each block.}
    \label{fig:resnet}
\end{figure}

\begin{table}[!b]
\vspace{-0.5em}
    \centering
    \scalebox{.68}{
    \addtolength{\tabcolsep}{0pt}
    \begin{tabular}{|l| c| c| c|} 
    \hline
    Method &  Error(\%) & Params Pruned(\%) & Pruned FLOPs(\%) \\ [0.8ex] 
    \hline\hline
    Li-pruned \cite{weightSumICLR-17} &  6.60 & 64.0 &34.20\\
    SBP \cite{neklyudov2017structured} &  7.50 & -- & 56.52\\
    AFP-E \cite{ding2018auto} &  7.06 & 93.3 & 79.69\\
    AFP-F \cite{ding2018auto} & 7.13 & 93.5 & 81.39\\
    
    \textbf{PP-1 (Ours)} & \textbf{6.54}&  92.5 & \textbf{82.8} \\
    \textbf{PP-2 (Ours)} &  \textbf{6.65}&  \textbf{94.3} & \textbf{84.5} \\
    \hline
    \end{tabular}
    }
    \vspace{-5pt}
    \caption{Comparison of pruning VGG-16 on CIFAR-10 (the baseline accuracy is 93.49\%).}
    \label{tab:vgglike-16}
    \vspace{-1em}
\end{table}

\subsection{VGG-16 on CIFAR-10}
We experimented with the VGG-16 on the CIFAR-10 dataset. We follow the same parameter settings and training schedule as \cite{weightSumICLR-17}. Table~\ref{tabvgglike} shows the layer-wise pruning statistics for PP-1 (first pruned model), and PP-2 (second pruned model).
We compare our results with the recent works on filter pruning. Our approach consistently performs better as compared to Li-pruned \cite{weightSumICLR-17}, SBP \cite{neklyudov2017structured}, AFP \cite{ding2018auto} as shown in  Table~\ref{tab:vgglike-16}.

\begin{table}[t]
    \centering
    \scalebox{0.8}{
    \addtolength{\tabcolsep}{9pt}
    \begin{tabular}{|c| c| c| l|} 
    \hline
    Alpha value & Error(\%) & Parameters & FLOPs \\ [0.8ex] 
    \hline\hline
    5 & 6.54 & $1.12\times10^6$ & $5.3\times10^7$\\
    10 & 6.54 & $1.13\times10^6$ & $5.4\times10^7$\\
    20 & 6.56 & $1.15\times10^6$ & $5.5\times10^7$ \\
    30 & 6.61 & $1.27\times10^6$ & $6.1\times10^7$ \\
    \hline
    \end{tabular}
    }
    \vspace{-5pt}
    \caption{Ablation study over the $\alpha$ values. Experimentally we found that $\alpha=10$ is the most suitable.}
    \label{tab:vgglike-16ablation}
\end{table}

\subsection{Ablation Study for VGG-16 on CIFAR-10}

This section shows a detailed analysis of the effect on the different component in the proposed approach. 
\subsubsection{Ablation Study on the hyper-parameter $\alpha$}
We did an ablation study on the hyper-parameter $\alpha$, i.e., how many filters are selected for partitioned $\mathcal{U}$. We experimented with $\alpha=5,10,20,30\%$. We found that if we take the lower value, it will not degrade the performance only it takes more epochs to prune. While if we take high $\alpha$ (say 30\%) value, it starts degrading the performance of model early, hence we are unable to get high pruning rate. In our case, we find $\alpha=10\%$ is a moderate value and at this rate, we can achieve a high pruning rate. This rate we set across all architecture like ResNet-50, VGG. 
\subsubsection{Pruning iterations Vs Error}
\begin{figure}
    \centering
    \includegraphics[scale=0.47]{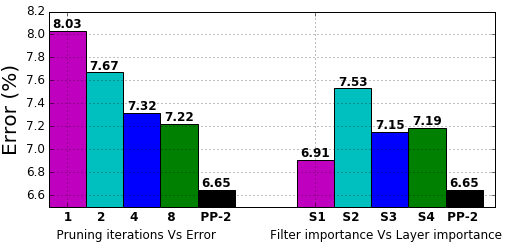}
    \caption{(a) Left figure shows the effectiveness of iterative pruning (b) Right figure shows effect of layer importance on error for the same FLOPs pruning (84.5\%)}
    \label{fig:ablation1}
    \vspace{-15pt}
\end{figure}

 In Fig.~\ref{fig:ablation1} (left) we have shown that if we do the pruning in 1 shot, it has significant accuracy drop (8.03\%) as compared to PP-2 on the same FLOPs pruning (84.5\%). While if we prune the model iteratively, we have less error rate for the same FLOPs pruning.

\subsubsection{Filter importance Vs Layer importance}
Most of the previous approach \cite{ding2018auto,weightSumICLR-17,yu2017nisp} focus on the \textit{How to prune} (filter importance/ranking), while it is also important for the better pruning rate to decide \textit{Where to prune}. Our approach also decides the \textit{where to prune} by considering layer importance. In the figure-\ref{fig:ablation1} (right) we are showing the ablation on our approach's capability to decide the layers importance. In figure-\ref{fig:ablation1} (right) we are showing the error rate on the similar FLOPs pruning without/with considering layer importance for the four compressed model search (S1,S2,S3,S4) with our approach (PP-2) using same filter importance criteria.

\begin{table}[!b]
    \centering
    \scalebox{0.8}{
    \addtolength{\tabcolsep}{6pt}
    \begin{tabular}{|l| c| c| c|} 
    \hline
    Method &  Error(\%) &  Pruned FLOPs(\%) \\ [0.8ex] 
    \hline\hline
    Li-B \cite{weightSumICLR-17} &  6.94 &  27.6\\
    NISP \cite{yu2017nisp} &  6.99 &  43.6\\
    CP \cite{channelPruning17} &  8.20 &  50.0\\
    SFP \cite{he2018soft} &  6.65 & 52.6\\
    AFP-G \cite{ding2018auto} &  7.06 &  60.9\\
    
    \textbf{PP-1 (Ours)}  & \textbf{6.91} &  \textbf{68.4}\\
    \hline
    \end{tabular} }
    \vspace{-5pt}
    \caption{Comparison of pruning ResNet-56 on CIFAR-10 (the baseline accuracy is 93.1\%).}
    \label{tab:resnet56}
\end{table}

\subsection{ResNet-56 on CIFAR-10}

We follow the same parameter settings and training schedule as \cite{weightSumICLR-17,he2018soft}. Our approach significantly outperforms various state-of-the-art approaches for ResNet-56 on CIFAR-10. The results are shown in Table~\ref{tab:resnet56}. We achieve high pruning rate $68.4\%$ with the $6.91 \%$  error rate, while AFP-G \cite{ding2018auto} has the error rate of $7.06 \%$ with only $60.9\%$  pruning.

\subsection{VGG-16 On ILSVRC-2012}

To show the effectiveness of our proposed approach, we also experimented with the large-scale dataset ILSVRC-2012 \cite{imagenet2015}. It contains 1000 classes with 1.5 million images. To make our validation set, randomly 10 images (from the training set) are selected from each class. This is used by PRC to calculate validation accuracy drop for adjusting the pruning rate. In this experiment, $\alpha$ is the same as the previous experiment. We follow the same setup and settings as \cite{channelPruning17}.

Our large-scale experiment for VGG-16 \cite{vgg2014very} on the ImageNet \cite{imagenet2015} shows the state-of-art result over the other approaches for model compression. Channel-pruning (CP) \cite{channelPruning17} has the $80.0\%$ model FLOPs compression with the top-5 accuracy $88.2\%$, while we have same FLOPs pruning (80.2\%) with the top-5 accuracy $89.81\%$. Refer to table-\ref{tab:vgg16Imagenet} for the detail comparison results. Our compressed model (PP-1) is obtained after 38 epochs.

\begin{table}[!t]
    \centering
    \scalebox{0.82}{
    \addtolength{\tabcolsep}{0pt}
    \begin{tabular}{|l| c| c| c| c|}
    \hline
    Method  & Top-5 Accu.(\%) & Pruned FLOPs(\%) \\ [0.1ex] 
    \hline\hline
    RNP (3X)\cite{lin2017runtime}  &  87.57 & 66.67\\
    ThiNet-70 \cite{luo2017thinet}  &  89.53 & 69.04\\
    CP \cite{channelPruning17}  &  88.20 & 80.00\\ 
    \textbf{PP-1 (Ours)}  & \textbf{89.81} & \textbf{80.20}\\
    \hline
    \end{tabular} }
    \vspace{-5pt}
    \caption{Comparison of pruning VGG-16 on ImageNet (the baseline accuracy is 90.0\%).}
    \label{tab:vgg16Imagenet}
    \vspace{-1em}
   \end{table}

\begin{table}[!t]
    \centering
    \scalebox{0.72}{
    \addtolength{\tabcolsep}{0pt}
    \begin{tabular}{|l|c|c|c|}
            \hline
            Method & Top-5 Accu.(\%)  & Parameters  & Pruned FLOPs(\%)\\[0.1ex] \hline\hline
            
           ThiNet \cite{luo2017thinet}      & 90.7 & 16.94M   & 36.9\\
           SFP \cite{he2018soft}   & 92.0   &  --   & 41.8\\
            
            \textbf{PP-1 (Ours)}  & \textbf{92.0} & \textbf{15.7M}  & \textbf{44.1}\\ \hline \hline
            CP \cite{channelPruning17}  & 90.8 & $ \sim $ 18M  & 50.0 \\ 
             \textbf{PP-2 (Ours)}  & \textbf{91.4} & \textbf{13.7M} & \textbf{52.2}\\ \hline
        \end{tabular}
    }
    \vspace{-5pt}
    \caption{Comparison of pruning ResNet-50 on ImageNet (the baseline accuracy is 92.2\%).}
    \label{resnettable}
    \vspace{-1em}
\end{table}

\begin{table}[!b]
\centering
\scalebox{.9}{
\addtolength{\tabcolsep}{1pt}
\begin{tabular}{|c|c|c|c|c|}
\hline
\multirow{2}{*}{\textbf{Model}}& \multirow{2}{*}{\textbf{data}} & \multicolumn{3}{c|}{\textbf{Avg. Precision, IoU:}}  \\ \cline{3-5}
 & & \textbf{0.5:0.95} & \textbf{0.5} & \textbf{0.75}  \\ \hline
\textbf{F-RCNN original} & trainval35K & 30.3 & 51.3 & 31.8   \\ \hline
\textbf{F-RCNN pruned} & trainval35K & 30.3 & 51.1 & 31.7  \\ \hline
\end{tabular}
}
\vspace{-5pt}
\caption{Generalization results on MS-COCO. Pruned ResNet-50 (PP-2) used as a base model for Faster-RCNN.}
\label{tab:coco}
\vspace{-1em}
\end{table}

\subsection{RESNET-50 ON ILSVRC-2012}

In ResNet, there exist restrictions on the few layers due to its identity mapping (skip connection). Since for $output=f(x)+x$ we need to sum two vector, therefore we need $x$ and $f(x)$ should have same dimension. Hence we cannot change the output dimension freely. Hence only two convolutional layers can be pruned for each block (see Fig-\ref{fig:resnet}). Unlike the previous work \cite{luo2017thinet} where they explicitly set p=q, we have not imposed any such restriction which results in more compression with better accuracy.
We prune ResNet-50 from block 2a to 5c continuously as described in the proposed approach. If the filter is pruned, then the corresponding channels in the batch-normalization layer and all dependencies to that filter are also removed. We follow the same settings as \cite{channelPruning17}.

Our results on ResNet are shown in Table~\ref{resnettable}. We are iteratively pruning convolutional filters in each epoch as described earlier. PP-1 is obtained after 34 epochs. Similarly, PP-2 is obtained after 62 epochs. We have experimentally shown that our approach reduces FLOPs and Parameters without any significant drop in accuracy.

\subsection{Practical Speedup}

The practical speedup is sometimes very different by the result reported in terms of FLOPs prune percentage. The practical speedup depends on the many other factors, for example, intermediate layers bottleneck, availability of data (batch size) and the number of CPU/GPU cores available. 
 
\begin{figure}[!t]
\vspace{-1em}
    \centering
    \includegraphics[scale=0.48]{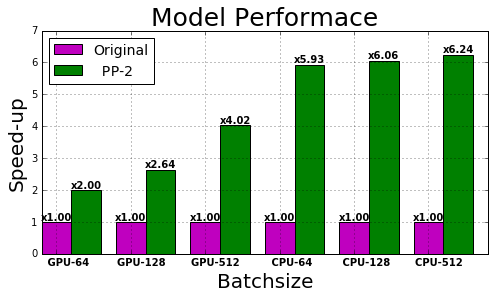}
    \caption{Speedup corresponding to CPU (i7-4770 CPU@3.40GHz) and GPU (GTX-1080) over the different batch size for VGG-16 on CIFAR-10.}
    \label{fig:speedupcpugpu}
    \vspace{-1em}
\end{figure}

For VGG-16 architecture with the 512 batch size, we have 4.02X practical GPU speedup, while the theoretical speedup is 6.43X (figure-\ref{fig:speedupcpugpu}). This gap is very close on the CPU, and our approach gives the 6.24X practical CPU speedup compare to 6.43X theoretical (Fig.~\ref{fig:speedupcpugpu}).

\subsection{Generalization Ability}
To show the generalization ability of our proposed approach, we also experiment on the object detection architecture. In this experiment we have taken the most popular architecture Faster-RCNN \cite{ren2015fasterrcnn} on MS-COCO \cite{lin2014coco} dataset.
\subsubsection{Compression for Object Detection}
The experiments are performed on COCO detection
datasets with 80 object categories \cite{lin2014coco}. Here all 80k
train images and a 35k val images are used for training (trainval35K) \cite{lin2017feature}. We are reporting the detection accuracies over the 5k unused val images (minival). In this first, we trained Faster-RCNN with the ImageNet pre-trained ResNet-50 base model. The results are shown in table-\ref{tab:coco}. 
In this experiment we used our pruned ResNet-50 model (PP-2) as given in Table-\ref{resnettable} as a base network in Faster-RCNN. We found that the pruned model shows similar performances in all cases. In the Faster-RCNN implementation, we use ROI Align and use stride 1 for the last block of the convolutional layer (layer4) in the base network. 
\section{Conclusion}
We proposed a Play and Prun Filter Pruning (PP) framework to prune CNNs. Our approach follows a min-max game between two modules (AFP and PRC). Since our approach can prune the entire convolution filter, there is a significant reduction in FLOPs and the number of model parameters.
Our approach does not require any special hardware/software support, is generic, and practically usable. We evaluated it on various architectures like VGG and Resnet. Our approach can also be used in conjunction with pruning methods such as binary/quantized weights, weight pruning, etc. These can directly be applied to the pruned model given by our method to get further speedups and model compression. The experimental results show that our proposed framework achieves state-of-art results on ResNet and VGG architecture and generalizes well for object detection task.


\newpage
\bibliographystyle{named}
\bibliography{ijcai19}

\end{document}